\documentclass{article}
\pdfoutput=1
\usepackage[final]{corl_2020} 

\usepackage[utf8]{inputenc} 
\usepackage[T1]{fontenc}    
\usepackage{hyperref}       
\usepackage{url}            
\usepackage{booktabs}       
\usepackage{amsfonts}       
\usepackage{nicefrac}       
\usepackage{microtype}      
\usepackage{color,xcolor}
\usepackage{epsfig}
\usepackage{graphicx}
\usepackage{algorithm,algorithmic}

\usepackage{adjustbox}
\usepackage{array}
\usepackage{booktabs}
\usepackage{colortbl}
\usepackage{float,wrapfig}
\usepackage{framed}
\usepackage{hhline}
\usepackage{multirow}
\usepackage{subcaption} 
\usepackage[font=small]{caption}
\usepackage[percent]{overpic}

\usepackage{amsmath,amsfonts,amssymb}
\usepackage{amsthm} 
\usepackage{bm}
\usepackage{nicefrac}
\usepackage{microtype}
\usepackage{courier}

\usepackage{changepage}
\usepackage{extramarks}
\usepackage{fancyhdr}
\usepackage{lastpage}
\usepackage{setspace}
\usepackage{soul}
\usepackage{xspace}
\usepackage{cuted}
\usepackage{fancybox}
\usepackage{afterpage}
\usepackage[subtle]{savetrees}

\usepackage{enumerate}
\usepackage{paralist,tabularx}
\usepackage{comment}
\usepackage{pdfpages}

\usepackage[font=small]{caption}

\makeatletter
\DeclareRobustCommand\onedot{\futurelet\@let@token\@onedot}
\def\@onedot{\ifx\@let@token.\else.\null\fi\xspace}

\def\eg{e.g\onedot, } 
\def\ie{i.e\onedot, }

\def\etal{et al\onedot}
\makeatother

\definecolor{MyDarkBlue}{rgb}{0,0.08,1}
\definecolor{MyDarkGreen}{rgb}{0.02,0.6,0.02}
\definecolor{MyDarkRed}{rgb}{0.8,0.02,0.02}
\definecolor{MyDarkOrange}{rgb}{0.40,0.2,0.02}
\definecolor{MyPurple}{RGB}{111,0,255}
\definecolor{MyRed}{rgb}{1.0,0.0,0.0}
\definecolor{MyGold}{rgb}{0.75,0.6,0.12}
\definecolor{MyDarkgray}{rgb}{0.66, 0.66, 0.66}

\newcommand{\mypara}{\vspace{0mm}\noindent\textbf}

\newcommand{\OURS}{Fit2Form\xspace}
\newcommand{\degree}{$^{\circ}$}

\newcommand\blfootnote[1]{%
  \begingroup
  \renewcommand\thefootnote{}\footnote{#1}%
  \addtocounter{footnote}{-1}%
  \endgroup
}

\title{Fit2Form: 3D Generative Model for Robot Gripper Form Design }

\author{
  Huy Ha$^*$, \quad Shubham Agrawal$^*$,
  \quad Shuran Song \\ 
  Columbia University, New York, NY, United States \\
  \url{https://fit2form.cs.columbia.edu/}
  \vspace{-6mm}
}

\begin{document}
\maketitle

\begin{abstract}
The 3D shape of a robot's end-effector plays a critical role in determining it's functionality and overall performance. 
Many industrial applications rely on task-specific gripper designs to ensure the system's robustness and accuracy.  
However, the process of manual hardware design is both costly and time-consuming, and the quality of the resulting design is dependent on the engineer's experience and domain expertise, which can easily be out-dated or inaccurate.
The goal of this work is to use machine learning algorithms to automate the design of  task-specific gripper fingers.
We propose Fit2Form, a 3D generative design framework that generates pairs of finger shapes to maximize design objectives (i.e., grasp success, stability, and robustness) for target grasp objects.
We model the design objectives by training a Fitness network to predict their values for pairs of gripper fingers and their corresponding grasp objects.
This Fitness network then provides supervision to a 3D Generative network that produces a pair of 3D finger geometries for the target grasp object.
Our experiments demonstrate that the proposed 3D generative design framework generates parallel jaw gripper finger shapes that achieve more stable and robust grasps compared to other general-purpose and task-specific gripper design algorithms.
Code and data links can be found at \href{https://fit2form.cs.columbia.edu/}{https://fit2form.cs.columbia.edu/}.
\end{abstract}
\vspace{-3mm}
\blfootnote{$*$ Indicates equal contribution}

\vspace{-2mm}
\section{Introduction}
\begin{figure}[t]
    \centering
    \includegraphics[width=\linewidth]{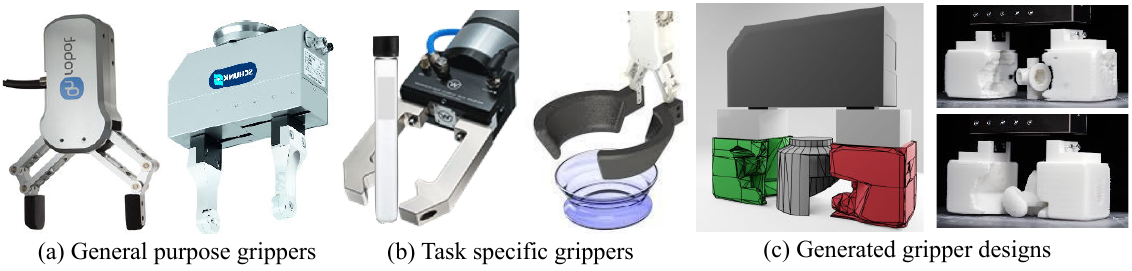}
    \caption{\textbf{Robot Gripper Design.}
    While most of the recent works have been focused on learning robust control policy for general-purpose grippers (a), the majority of robot grippers in industrial applications are highly customized to improve the system's robustness and accuracy for the target task (b). However, the process of manual hardware design is costly and time-consuming. The goal of \OURS is to automate this design process with a data-driven approach and generate gripper geometry for a target object that would satisfy the design objectives (c). \label{fig:example}}
    \vspace{-5mm}
    
\end{figure}

\vspace{-2mm}
As the physical interface between a robot and its environment, the end-effector and its design plays a critical role in determining the system’s functionality and performance \cite{rodriguez2013effector}. 
To grasp different objects, most of the recent works  \cite{caldera2018review,mahler2017dex,zeng2017robotic, Zeng2018a} focused on learning sophisticated control policies for a generic gripper (Fig.\ref{fig:example} a) instead of designing new customized grippers.
While these prior works have significantly improved control algorithms for dexterous manipulation, the application of data-driven approaches on gripper hardware design is still under-explored.
Can a system learn to generate customized gripper fingers for different grasp objects and design objectives?

Even though task-specific grippers are commonplace in real-world industrial applications (e.g: Fig.\ref{fig:example} b), the manual design process depends on many costly trial-and-error experiments and the engineers' experience, which may be outdated or inaccurate.
This prevents fast adaptations to changes in manufacturing processes that are frequent in small and medium-sized enterprises.

The goal of this paper is to automate the robotic gripper geometry design process with machine learning: given a target grasp object, the algorithm generates the 3D geometry of a pair of fingers for a parallel jaw gripper that maximizes the design goals (\eg grasp success, stability, and robustness). 
There are two main challenges in achieving this goal: 
(1) How to represent a set of design objectives in a differentiable manner to enable efficient learning, and (2) How to effectively navigate a large design space to produce a valid solution? 

First, to effectively model the design objectives, we train a Fitness network to predict the values of design objectives (or fitness score) for a given pair of gripper fingers and a target object.  
The Fitness network combines high-level design objectives (\eg grasp success, stability, robustness, etc.) into a single differentiable objective function.
The parameters of the Fitness network are trained in a self-supervised manner via simulating grasps on random fingers and objects.

Second, to efficiently navigate the large design space (\ie all possible 3D shapes), we use a 3D Generator network that models a general distribution of 3D shapes and produces 3D finger geometries based on the target grasp object.
The parameters of the Generator network can be directly optimized to maximize the Fitness network's fitness score predictions.
During training, both the Generator network and the Fitness network are optimized jointly in the co-training phase.

In summary, our primary contribution is a novel 3D Generative Design framework, \textbf{\OURS}, that leverages data-driven algorithms to automate the design process of robot hardware. 
Our experiments demonstrate that \OURS can generate finger shapes for parallel jaw gripper that achieve more stable and robust grasps given a target object compared to other general-purpose (e.g., WSG50) and task-specific grippers (e.g., imprint-based designs).

\vspace{-1mm}
\section{Related work}
\vspace{-1mm}
Prior works have investigated manual design and engineering \cite{causey1999elements}, and kinematic structure optimization \cite{brown19993} of robotic gripper end-effectors. Here, we primarily review works that consider geometry optimization for 3D design.

\mypara{Geometry Optimization for Robot Hardware Design:}
Geometry optimization has been an important part of mechanical design, with applications in designing gears  \cite{tong1998generation,liu2008design}, part feeders  \cite{caine1994design} and fences \cite{brokowski1995optimal}. 
For automatic finger design, Wolniakowski \etal \cite{wolniakowski2015task} optimized the parameterized geometry of simple cuboid based gripper using various gripper quality metrics computed in simulation. 
To improve the throwing distance, Taylor and Rodriguez \cite{taylor2019optimal} optimized the curvature of the end-effector and motion trajectory. 
However, finding an effective parameterization is a tedious process requiring domain expertise, which often results in a severely limited design space. 
In contrast, our model is able to efficiently explore the full 3D geometric space by representing 3D geometry with Truncated Signed Distance Functions (TSDF) \citep{newcombe2011kinectfusion}.

Topology optimization \cite{bendsoe2013topology,wu2019design} optimizes geometry to reduce material usage while maintaining the strength.
However, these methods can only improve upon pre-existing designs, thus still requires manual engineering while being prone to local optima.
Additionally, its optimization only considers two objectives, while our algorithm is able to explore a larger unconstrained design space without human intervention to maximize any number of design objectives.

Heuristic imprinting  \cite{velasco1998computer, schwartz2017designing, velasco1996approach} is another popular method for generating fingers with complementary geometry to its grasp objects, thus producing grippers which cage the grasp objects. 
Our experiments show that these fingers require lengthy manual tuning phases to work properly and are not robust to perturbations in initial object orientation.
However, we leverage imprint fingers to efficiently guide our network's exploration of the design space, while exposing the network to a larger space of finger geometries to encourage diversity and improvement beyond imprints (Sec. \ref{section_approach}).
%

%
%

\mypara{Automatic Design with Evolutionary Algorithms:}
Lipson \etal \cite{cheney2014unshackling} proposed a method for automatic design and manufacturing of soft robots using evolutionary algorithms \cite{lipson2000automatic}, which simultaneously explores the space of possible morphological structures, actuation mechanisms, and physical material properties to maximize its walking speed. 
However, evolutionary algorithms depend on simulations for the optimization signal even at test time, and so are slow at generating new instances.
Our proposed method uses a Fitness network as a fast differentiable proxy for simulation, and can generate new fingers with a forward pass through the Generator network.


\mypara{Grasp Quality based Optimization:} Most literature in robot grasping have used analytical \cite{Prattichizzo2008, miller2004graspit, rodriguez2012caging} or learned grasp quality metrics \cite{lenz2015deep, detry2013learning, herzog2014learning} to select and evaluate the grasping pose. However, as shown by \cite{goins2016implementation}, the grasp quality computed by analytical methods often only models static contact points without considering the dynamic grasping process and environment constraints and hence does not translate well to the real world. Instead, we use a physics simulation environment to simulate the complete dynamic grasping process. Most related to our approach is GQN from Dexnet-2.0 \cite{mahler2017dex}, where grasp quality is predicted by a neural network. However, their approach assumes a fixed gripper shape. Our Fitness network takes into account both the object and gripper finger geometries, and provide values for design objectives for the pair. 

\mypara{Generative Models for Design:}
Many recent works have leveraged powerful generative models in creative design, such as image manipulation \cite{zhu2016generative,bau2018gan} or sketch generation \cite{jin2017towards, Liu_2019_CVPR} which use 2D generative networks \cite{Goodfellow2014Generative} to model the distribution of the target domain (\eg  paintings or photos), and then use learned distribution to facilitate the design process.
 3D generative models have been developed for tasks such as shape completion \cite{3DShapeNets}, classification \cite{3DShapeNets,Wu2016Learning}, and editing \cite{liu2017interactive}.  However, these 3D generative frameworks only focus on modeling objects' shape distribution based on objects' semantic categories and do not generate 3D geometry based on given design objectives. In this work, we leverage an additional differentiable Fitness network to guide the 3D Generative network in generating desired 3D shapes.
 


\vspace{-2mm}
\section{Approach}\label{section_approach} \vspace{-2mm}
\begin{figure}[t]
    \centering
    \vspace{-3mm}
    \includegraphics[width=\linewidth]{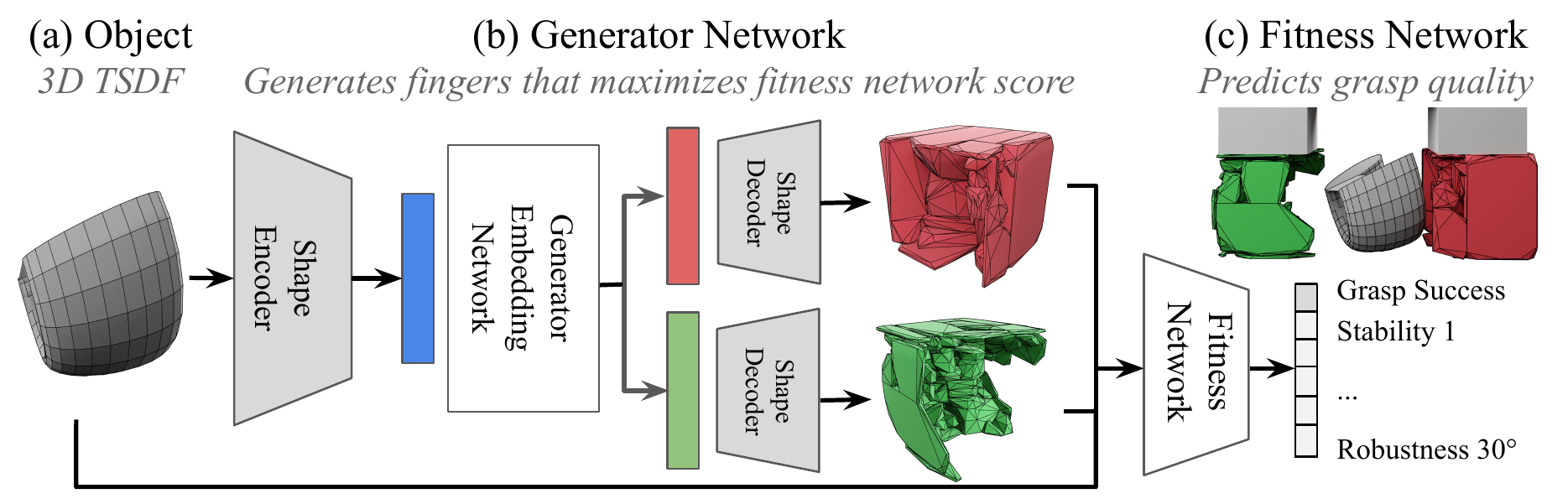}
    \caption{\textbf{\OURS Overview.}  Our 3D generative design framework consists of two major components (1) a Generator network which efficiently navigates the unconstrained 3D design space to produce finger volumes given the object, and (2) a Fitness network which predicts values for design objectives (\eg grasp success, stability, and robustness) given the object, left and right finger.
    The Generator is trained to maximize all predicted scores from the Fitness network.  
    During co-training, both the Generator network and Fitness network are optimized jointly. 
    }
    \label{fig:architecture}
    \vspace{-5mm}
\end{figure}

We formulate the gripper finger geometry design problem as follows: 
for a given target object $o$, the  goal of a gripper designer is to generate 3D shapes of left $f_l$ and right $f_r$ finger for a parallel jaw gripper, where generated fingers should simultaneously maximize $m$ objectives $\mathcal{T}_1,\mathcal{T}_2,\dots, \mathcal{T}_m$. 
We focus on three types of objectives, i.e., (1) \ul{grasp success} $\in \{0,1\}$ whether the generated fingers successfully grasped the target object, (2) \ul{force stability} $\in \{0, 1\}$ whether the grasp was maintained after application of certain forces for a certain duration, (3) \ul{robustness} $\in \{0, 1\}$ whether the generated fingers grasped the object with perturbations in orientation.

To achieve this goal, \OURS have two main components: a Generator network $\mathcal{G}$ which produces finger volumes given the grasp object, and a Fitness network $\mathcal{F}$ which predicts values for design objectives given the object $o$, left finger $f_l$, right finger $f_r$. We provide details on how we evaluate the design objectives in Sec. \ref{subsection_sim}, followed by descriptions of the Fitness and Generator network in Sec.  \ref{subsection_fn}  and \ref{subsection_gn}. Lastly,  we will discuss the co-training of Fitness and Generator network in Sec. \ref{subsection_train}.

\vspace{-1mm}\subsection{Grasp Simulation} \vspace{-1mm}
\label{subsection_sim}
For all experiments in this paper, we use the Schunk WSG50 parallel jaw gripper as the base.
This gripper base's modular design allows us to quickly swap out our 3D printed fingers using a custom mount (Fig. \ref{fig:example}c, Fig. \ref{fig:gripper}).
We define a cubical bound of side lengths $6 \mathrm{cm}$ that a generated finger can take (Fig. \ref{fig:data_pipeline} b).
Apart from the geometry of the object being grasped, finger design should also consider the gripper strength, kinematics, and dexterity of the control algorithm.
However, in this work, we focus on finger geometry design and hence, utilize a fixed top-down grasp policy (i.e.,  the gripper will always attempt to grasp at the same position and orientation).

We generate our target grasping objects using the ShapeNet dataset \cite{shapenet2015}.
For each target object, we sample a 3D object from ShapeNet and allow it to stabilize from a random orientation on the ground before re-scaling it to a cubical view bound of side lengths $6 \mathrm{cm}$, centered at the grasp location.
The re-scaled object is then added to a dataset used for grasping called the \ul{target object dataset}.
Thus, objects from the target object dataset are stable before grasping and fit inside our network's view bounds.
All target objects have a mass of 50 grams, lateral friction of 0.2, and rolling friction of 0.001, while fingers have lateral friction 1, and rolling friction 1.

The design objectives are evaluated in the PyBullet physics simulation  \cite{coumans2019}.
To evaluate the grasp quality, we sample and load an object $o$ from the target object dataset and a WSG50 with our generated fingers ($f_l$,$f_r$). 
The geometries of $o$, $f_l$, and $f_r$ are encoded as $40 \times 40 \times 40$ volumetric TSDFs, and are converted to meshes for use in simulation with marching cubes \cite{lorensen1987marching} and approximate convex decomposition \cite{mamou2016volumetric}.

\begin{wrapfigure}{r}{0.48\textwidth}
    \centering
    \vspace{-4mm}
    \includegraphics[width=0.98\linewidth]{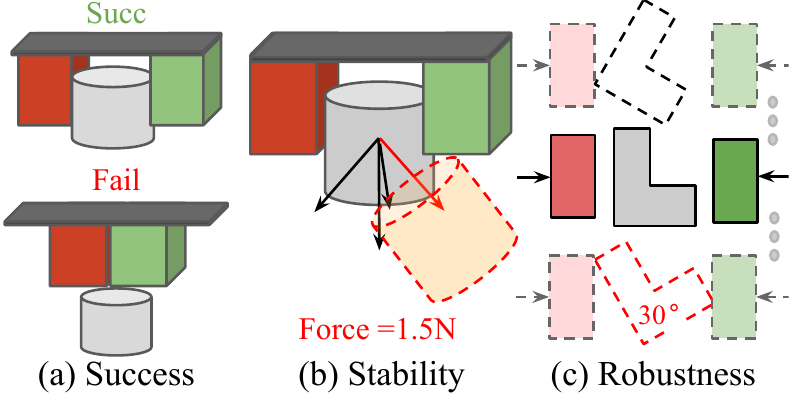} \vspace{-2mm}
    \caption{ \textbf{Design Objectives} are defined by a set of grasp quality measurements: (a) grasp success, (b) four force stability directions, and (c) different pre-grasp orientation of the target object for robustness. } 
    \label{fig:objective}
    \vspace{-3mm}
\end{wrapfigure}

We simulate the grasping process by closing the gripper centered around the target object with 1.0 N force, then lifting up for 30 cm, before checking the grasp success. To evaluate grasp stability, we sequentially apply a 1.5 N force for 20 time steps in 4 pre-defined directions: one along the negative z-axis (down direction) and the other three along the edges of a regular tetrahedron (see Fig \ref{fig:objective}). 
After the application of each force, we restore the post-grasp state of the system. %
Whether or not the grasp was maintained after the application of each force marks whether the stability test passed.
To evaluate the grasp robustness, we test the grasp success with the target object rotated with -30\degree,-20\degree, -10\degree,10\degree,20\degree, 30\degree along z-axis (up direction) (Fig \ref{fig:objective} c). %
The final grasp quality vector $\mathbf{\mathcal{T}}$ have dimension $m = 1 + 4 + 6 = 11$.

Apart from the grasp quality, we also want to ensure that the generated fingers can be manufactured (i.e., 3D printed).
To this end, \ul{feasible} fingers must be (1) significantly base connected, where at least 10\% of the finger base area is connected to the mount, and (2) one single connected component. If the generated finger shape is not fully connected, we use the largest connected component as the final shape. Otherwise, infeasible fingers receive a grasp score of zero for all design objectives.

\vspace{-1mm}
\subsection{Modeling Design Objectives with Fitness Network}\label{subsection_fn} \vspace{-1mm}

Given the TSDFs of $(o, f_r, f_l)$, the Fitness network predicts the values for ground truth objectives $\mathcal{T}_i(o, f_r, f_l)$ computed by the physics simulation. A successfully trained Fitness network acts as a fast and differentiable proxy for these non-differentiable design objectives.
To construct the input volume, we concatenate the input tuple $(o, f_r, f_l)$ side-ways along the grasping direction in order $(f_l, o, f_r)$, resulting in a $120 \times 40  \times 40$ volume. 
This concatenation helps the network compare local geometries of objects and fingers by positioning the volumes as they would appear spatially in a grasp.

The Fitness network contains 3D convolution layers with residual connections \cite{he2016deep} for down-sampling followed by fully-connected layers. 
Each layer is followed by batch normalization and leaky ReLU activations except the last layer which uses Sigmoid activation.
The last layer has $m$ nodes each corresponding to one design objective $\mathcal{T}_m$.
We use the Binary Cross Entropy loss between the estimated fitness score and the ground truth score $\mathcal{T}(o, f_l, f_r)$ obtained from the simulation:
\vspace{-1mm}
\begin{equation}
\small  
\label{eq:fitness_loss}
\mathcal{L}_\mathrm{fit} = 
\dfrac{1}{m}\sum_{i = 1}^m \mathrm{BCE}\bigg(\mathcal{F}_i(o, f_l, f_r), \mathcal{T}_i \bigg)
\end{equation}

\vspace{-2mm}
\mypara{Pretraining Fitness Network:}
Training the Fitness network requires a grasp dataset with datapoints of input volumetric TSDFs $(o,f_l,f_r)$ and output simulation grasp scores $\mathbf{\mathcal{T}}(o,f_l,f_r) \in \{0,1\}^m$.
To meet the data requirement of large neural networks, we generate a pretraining grasp dataset containing two types of fingers and their corresponding grasp scores on target objects:

\ul{Imprint fingers:}
To generate imprint fingers for a given object, a TSDF volume of the object using only two views along the grasp direction is negated and split along the middle. 
The resulting volumes are then passed to marching cubes to generate left and right imprint fingers (Fig. \ref{fig:data_pipeline}c). 
Using only two views ensures that the resulting fingers can close properly (see supplementary material for more details).
Imprint fingers are good task-specific gripper baselines and are popular in industry \cite{schunkegrip} and academia \cite{velasco1998computer,schwartz2017designing}. 

\ul{ShapeNet fingers:}
To encourage our networks to explore the large design space, we use objects from ShapeNet \cite{shapenet2015} to increase the diversity of the grasp dataset.
Since the fingers have to be significantly connected to the base, we search for such a plane with a significant base area within the top one-third volume of the finger and if found, slice and stretch it to the finger bounds (Fig. \ref{fig:data_pipeline}d).

\begin{figure}[t]
    \centering

    \includegraphics[width=\linewidth]{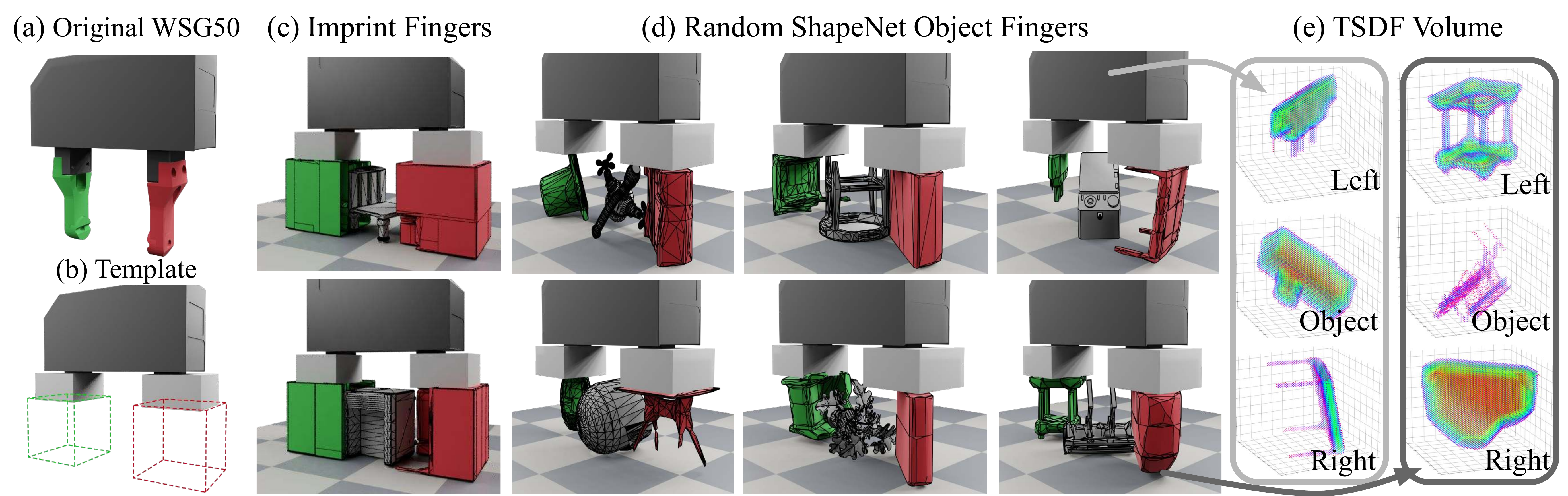}
    \caption{\textbf{Grasping Data Generation.} (a) Schunk WSG50 gripper with original fingers.  (b) Template of custom mount and finger volume bounds used to generate new grippers. (c) Imprint gripper (d) Random ShapeNet objects used as fingers. (e) 3D TSDF volume used by the network.}
    \label{fig:data_pipeline}
    \vspace{-6mm}
\end{figure}

\vspace{-1mm}
\subsection{Efficiently Exploring Design Space with Generator Network}\label{subsection_gn}\vspace{-1mm}
Efficiently navigating through a large space of finger designs to find feasible fingers with a good fitness score is a challenging problem.
Our experiments show that naively optimizing random embeddings for getting good fitness scores performs poorly (Tab.~\ref{tab:grasp_results}). 
Intuitively, for a particular object, we want fingers that cage the object like imprint fingers but at the same time are robust to perturbations in the object orientation. 
Hence, having a good Generator network will act as prior knowledge about what fingers might be good.
From each target object $o$'s TSDF, our Generator network generates a pair of 3D finger TSDFs $f_l$ and $f_r$ that maximizes the design objectives and satisfy the manufacturing constraints.

The Generator network $\mathcal{G}$ has three components: (1) a 3D shape encoder $\mathcal{E}_{\mathrm{shape}}$ that encodes the target objects' 3D TSDF volume into a 686-dimensional vector embedding, (2) an embedding network $\mathcal{G}_{\mathrm{emb}}$ that predicts two-finger shape embedding given a target object embedding, and (3) a shape decoder network $\mathcal{D}_{\mathrm{shape}}$ that decodes the fingers' TSDF volume from the embeddings produced by  $\mathcal{G}_{\mathrm{emb}}$. 
The encoder and decoder each consist of nine 3D convolution layers for down- and up-sampling with interleaving batch normalization layers and leaky ReLU activation functions.
The embedding network consists of three fully-connected layers with leaky ReLU activation functions. 
In the following paragraphs, we describe how we pretrain the shape encoder, decoder, and embedding network. In Sec.\ref{subsection_train}, we describe how to jointly optimize the $\mathcal{G}$ with supervision provided by the $\mathcal{F}$. 

\mypara{Learning Shape Distributions with 3D Auto-Encoder.} \label{subsection_vae} 
The goal of the $\mathcal{E}_{\mathrm{shape}}$ and $\mathcal{D}_{\mathrm{shape}}$ is to reduce $\mathcal{G}_{\mathrm{emb}}$'s search space to a low-dimensional $[-1, 1]^{686}$ latent space which represents a high-dimensional $[-1, 1]^{40 \times 40 \times 40}$ volumetric space.
We train a 3D auto-encoder with finger TSDF volumes sampled from the grasp dataset whose encoder and decoder are $\mathcal{E}_{\mathrm{shape}}$ and $\mathcal{D}_{\mathrm{shape}}$ respectively. 
Once trained, both $\mathcal{E}_{\mathrm{shape}}$ and $\mathcal{D}_{\mathrm{shape}}$ are then fixed for the rest of the process.
We observed that pretraining $\mathcal{E}_{\mathrm{shape}}$ and $\mathcal{D}_{\mathrm{shape}}$ greatly increases $\mathcal{G}$'s ability to generate feasible 3D shapes.

\mypara{Pretraining Generator Embedding Network:} 
To provide the Generator network a good initialization point for later exploration with Fitness network, the Generator network is pretrained to produce imprint fingers for a given input object with MSE loss.
Our experiment shows that this pre-training helps the Generator network with convergence during co-training. 

\vspace{-1mm}
\subsection{Generator and Fitness Network Co-Training} 
\label{subsection_train}
\vspace{-1mm}

After pre-training, the Generator network $\mathcal{G}$ and Fitness network $\mathcal{F}$ are co-trained for 20 cycles, where $\mathcal{G}$ tries to maximize $\mathcal{F}$'s fitness score prediction, and $\mathcal{F}$ tries to accurately predict grasp scores for a growing grasp dataset, consisting of both the pretrain grasp dataset and $\mathcal{G}$'s new co-train grasp dataset.

Each co-training cycle consists of three steps:
First, 512 target objects are sampled, and a new pair of fingers are generated using $\mathcal{G}$ for each object.
The ground truth design object values are computed using simulation, and these new data points are added to the co-train grasp dataset.
Second, we train $\mathcal{F}$ by sampling batches with 16 data points from the pretrain and co-train grasp dataset each.
Batches are also balanced for the success rate.
$\mathcal{F}$ is trained with $\mathcal{L}_\mathrm{fit}$ for at least 5 epochs, until it reaches a target loss of $0.2$ or for a maximum of $50$ epochs in the current cycle.
Finally, we train $\mathcal{G}$ using the target object dataset to maximize $\mathcal{F}$'s predicted fitness score, where $\mathcal{G}$'s loss is $\mathcal{L}_\mathrm{gen} =  -\mathcal{F}(o, \mathcal{G}(o)) $. 
This training proceeds for 5-50 epochs or a target loss $\mathcal{L}_\mathrm{gen}$ of $-0.9$ is achieved. Here, one epoch is limited to 500 optimization steps of batch size 32.

This co-training of $\mathcal{G}$ and $\mathcal{F}$ allows $\mathcal{F}$ to adapt its prediction to $\mathcal{G}$'s newly generated fingers, which might have a very different distribution from the pretrain grasp datasets, and better guide $\mathcal{G}$ to produce better fingers. The mixing of co-train grasp dataset with pretrain grasp datasets allows $\mathcal{F}$ to guide $\mathcal{G}$ out of local optima, beyond fingers that $\mathcal{G}$ is currently producing.
\begin{table}[t]
\vspace{-4mm}
\centering
\small
\setlength\tabcolsep{4pt}
\begin{tabular}{lccccccccccccc}
\toprule

& Grasp & \multicolumn{5}{c}{Stability}  & \multicolumn{7}{c}{Robustness} \\  
 \cmidrule(lr){3-7} \cmidrule(lr){8-14} 
&  Succ. & $S_1$ &  $S_2$ &  $S_3$ &  $S_4$ & avg. & -30\degree &  -20\degree &  -10\degree &  10\degree &  20\degree &  30\degree  & avg.  \\
\midrule
WSG50 & 25.8 & 3.3 & 2.1 & 2.1 & 1.9 & 2.4 & 12.2 & 14.3 & 17.6 & 17.5 & 14.6 & 12.9 & 14.8\\
Imprint \cite{schunkegrip} & 78.0 & 64.9 & 59.0 & 57.3 & 57.8 & 59.8 & 66.6 & 68.9 & 70.7 & 70.0 & 68.7 & 67.0 & 68.7 \\
\midrule
RandEmb+FN& 21.3 & 12.4 & 6.5 & 6.8 & 6.6 & 8.1 & 12.7 & 13.9 & 16.4 & 16.2 & 13.4 & 11.9 & 14.1 \\
ImpGN & 28.9 & 13.3 & 10.4 & 10.5 & 10.3 & 11.1 & 17.1 & 20.0 & 23.2 & 23.0 & 19.5 & 16.3 & 19.8\\
Fit2Form/ImpGN &81.8&65.6&54.8&59.0&56.6& 59.0 &72.0&73.9&76.2&75.1&74.2&73.4& 74.1\\
\midrule
Fit2Form GS & 85.2 & 64.3 & 57.0 & 58.6 & 53.3 & 58.3 & 71.0 & 75.2 & 78.5 & 78.0 & 74.0 & 69.7 & 74.4\\
Fit2Form GS + R & 81.8 & 62.8 & 51.9 & 53.3 & 50.9 & 54.7 & 70.6 & 72.9 & 75.5 & 74.6 & 71.9 & 69.5 & 72.5\\
Fit2Form GS + S & 87.5 & \textbf{70.8} & \textbf{66.0} & \textbf{62.6} & \textbf{64.1} & \textbf{65.9} & 75.7 & 78.1 & \textbf{81.4} & \textbf{81.2} & 78.3 & 75.3 & 78.3\\
Fit2Form & \textbf{87.6} & 69.0  & 61.6 & 60.5  & 61.7  & 63.2 & \textbf{76.1}  & \textbf{78.3} &81.1  & 81.1  & \textbf{79.2}  & \textbf{76.8} & \textbf{78.8}\\
\bottomrule
\end{tabular}
\vspace{1mm}
\caption{\textbf{Quantitative Evaluation.}}
\label{tab:grasp_results}
\vspace{-7mm}
\end{table}

\vspace{-2mm}
\section{Evaluation}
\vspace{-2mm}
\mypara{Data:} We trained Fit2Form on a target object dataset sampled from $39$ ShapeNet categories, and test them on  $8$ unseen ShapeNet object categories. We also include $13$ unseen adversarial objects from Dex-Net 2.0 \cite{mahler2017dex} in the test set. In total, we use 161,804 objects for training and 5,127 objects for evaluation.

\mypara{Metrics:}  As described in Sec \ref{section_approach}, we use three design objectives to evaluate the quality of generated fingers: grasp success, stability under various forces, and robustness to initial object orientation. For all experiments, we use a top-down grasp, centered around the target object with no rotation along the z-axis, with 1.0N grasping force.
 This small grasp force was chosen to increase the task difficulty and highlight differences between different gripper designs.

\mypara{Baselines:} We compare our algorithm with the following alternative gripper designs:
(1) [WSG50] the original fingers from the WSG50 gripper,
(2) [Imprint] the Imprint based grippers (described in Sec. \ref{subsection_fn}),
To analyze the effectiveness of the various parts of our approach, we also compare our method with: 
(3) [RandEmb+FN] Test time optimization of random finger embeddings using the shape decoder $\mathcal{D}_{\mathrm{shape}}$ and Fitness network without the Generator network,
(4) [ImpGN] the Generator pretrained with the imprint grasp dataset only without co-training with Fitness network,
(5) [Fit2Form/ImpGN] Our method without pretraining the Generator on the imprint grasp dataset. 

\begin{figure}[t]
    \vspace{-3mm}
    \includegraphics[width=\linewidth]{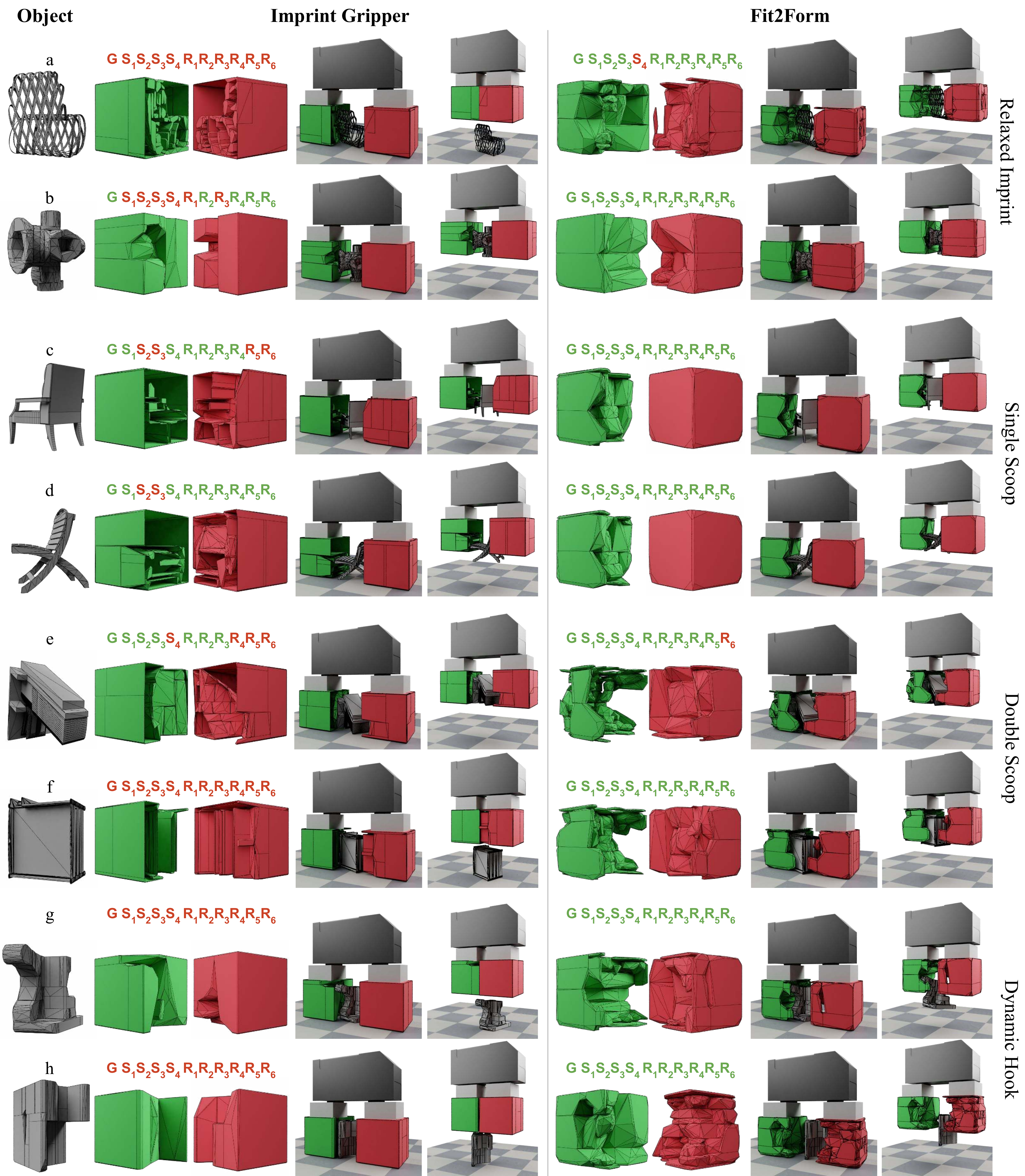}
    \caption{\textbf{Qualitative Results.} 
    We compare with grippers generated using the Imprint baseline and highlight four typical strategies used by Fit2Form: relaxed imprint (a,b), single-sided scoop (c,d), double-sided scoop (e,f), and dynamic hook (g,h).
    The green/red string above each gripper indicates success/failure of grasp success, 4 stability, and 6 robustness metrics.
    Objects in b,e,g,h adversarial objects from DexNet 2.0, the rest are from ShapeNet.
    }
    \vspace{-6mm}
    \label{fig:result}
\end{figure}
\vspace{-1mm}
\subsection{Experimental Results}
\vspace{-1mm}
\mypara{Comparison with WSG50 Gripper:}
The WSG50, which was designed for general purpose manipulation, has special geometry on its fingertips to increase friction and contact surface for better grasping. 
Tab.~\ref{tab:grasp_results} shows that WSG50 fingers performs much worse than Fit2Form fingers (-61.8\% in grasp success), highlighting the benefits of using task-specific gripper designs when the target object mass to grasp force ratio is high.

\mypara{Comparison with Imprint Gripper:}
Imprint grippers are designed to achieve perfect form closure for the target object which maximizes grasp stability. 
However, we observed that premature contact with target objects often prevent imprint fingers from properly closing around the objects, leading to misaligned and failed grasps (e.g., Fig.~\ref{fig:result} a,f,g,h).
Naturally, this sensitivity to object poses results in lower robustness scores compared to \OURS grippers (-10.1\% in robustness).

\mypara{Effects of using a Generator Network:}
In [RandEmb+FN], random finger embeddings are optimized for 200 steps to maximize Fitness network's grasp score prediction for a given $o$.
Despite having access to the same Fitness network, [RandEmb+FN] performs much worse ($-66.3 \%$ in grasp success) than \OURS's Generator network.
This demonstrates that a good Fitness network alone can lead to fingers stuck in local optima, while a Generator network trained with \OURS's pipeline can efficiently explore the unconstrained 3D design space to produce high performance and feasible fingers.

\mypara{Effect of co-training with Fitness Network and pretraining with Imprints:}
In this experiment, we investigate the effect of the Fitness network's supervision.
Without the co-training phase with the Fitness Network, [ImpGN] performs significantly worse than our approach ($-58.7\%$ in grasp success).
In contrast, a randomly initialized Generator Network with supervision from the Fitness network resulted in only a small drop in performance ($-5.8\%$ in grasp success).
Thus, while a good Generator Network initialization helps, the Fitness Network's supervision during the co-training phase was more crucial to the final performance.

\mypara{Effect of objective functions:}
We explore optimizing Fit2Form with only grasp success [Fit2Form GS], grasp success and robustness [Fit2Form GS + R], grasp success and stability [Fit2Form GS + S], and all objectives [Fit2Form] by evaluating their performance after 18 cotrained cycles.
Table~\ref{tab:grasp_results} show that stability is an important optimization metric, as removing it in [Fit2Form GS + R] results in $-8.5\%$ in stability and $-6.3\%$ in robustness compared to optimizing all metrics [Fit2Form].
In contrast, robustness is not required to perform well, as [Fit2Form GS + S]'s scores are close to [Fit2Form].

\mypara{What does Fit2Form learn?}
Fig. \ref{fig:result} shows examples of generated finger shapes for different target objects. By inspecting the qualitative results, we found that \OURS frequently use the following strategies to improve its grasping quality. 
\ul{Relaxed imprint finger}: the generated fingers form a closure around the object in a similar fashion to imprint fingers, with some extra space for better robustness. This strategy is particularly beneficial when the target object has a complicated thin structure (Fig. \ref{fig:result} a,b), where a precise imprint geometry can fail due to subtle misalignment.  
\ul{Single-sided scooping}: in many cases, one finger is generated with a simple flat surface and the other finger is generated with a cage-like shape. The flat surface pushes the object into the other fingers cavity and results in a more stable position (Fig. \ref{fig:result} c,d). 
\ul{Double-sided scooping}: similar to single-sided scooping but both fingers are generated with cage shapes, which the algorithm prefers over single-sided scooping for objects with larger volumes (Fig. \ref{fig:result} e,f).
\ul{Dynamic hook}: instead of directly creating a form closure around the target object, the gripper hooks target object's protruding structures during the lifting process (Fig. \ref{fig:result} g,h). Please see the supplementary material for additional results.
    
\begin{figure}[t]
    \centering

    \includegraphics[width=0.98\linewidth]{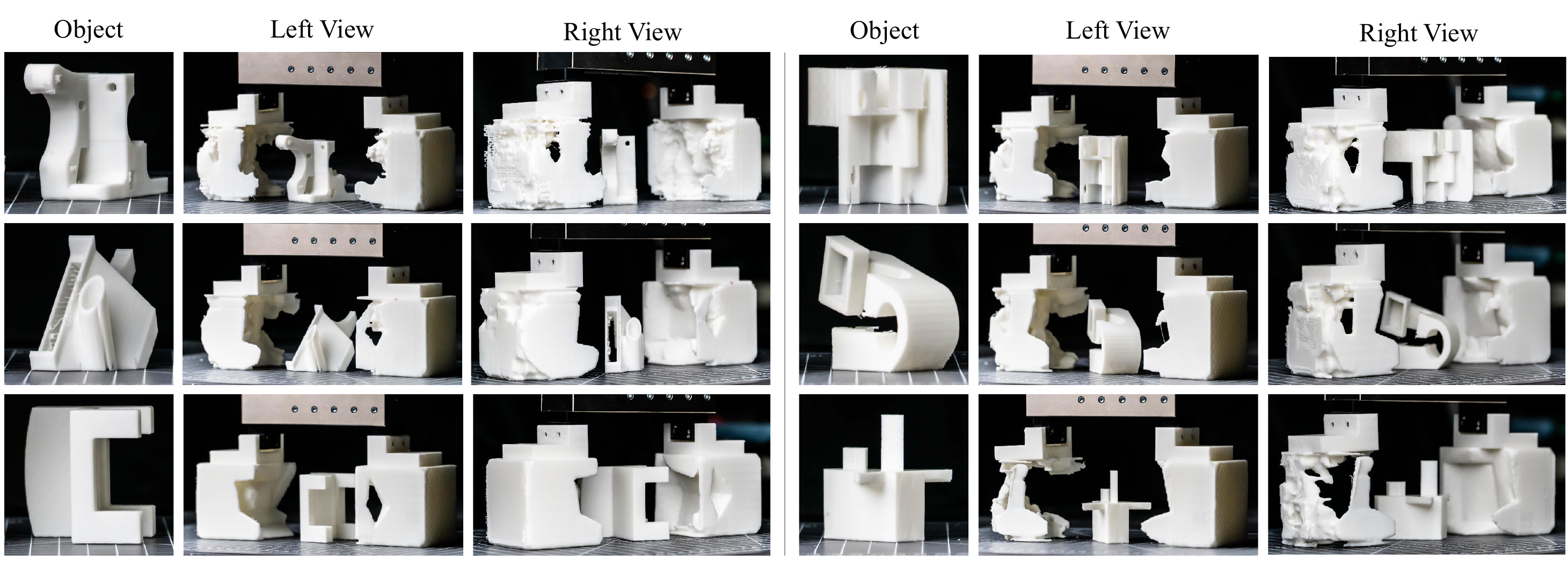} \vspace{-2mm}
    \caption{\textbf{3D printed Fit2Form gripper fingers}.
    We evaluated the manufactured Fit2Form finger's success, stability, and robustness, and observed that all 4 strategies discovered in simulation were also present in the real-world experiments.
    }
    \label{fig:gripper}
    \vspace{-5mm}
\end{figure}

\mypara{Manufacturing generated gripper.} 
To demonstrate feasibility, Fit2Form's gripper fingers were 3D printed along with their corresponding grasp objects (Fig.~\ref{fig:gripper}) using the Sindoh 3DWox 1.
For each set, we mount our gripper fingers onto the WSG50 gripper with a custom mount and position the grasp object into its rest pose with the lowered gripper as reference.
To evaluate grasp success, we execute a top-down grasp with 5N of force (the minimum force limit of the gripper) followed by a 15cm vertical lift.
Upon a successful grasp, we test for stability by shaking the gripper with the object along the X and Y axes, then poking the object with roughly 30N of force.
Finally, we test for robustness rotating the gripper along the z-axis by the specified angle before executing a grasp and measuring grasp success.
We observed that the 4 strategies Fit2Form discovered in simulation were also present in the real-world experiments.
Qualitative results for the grasp can be found at \href{https://fit2form.cs.columbia.edu/}{https://fit2form.cs.columbia.edu/}.

\vspace{-2mm}
\section{Conclusion and Future Directions} \vspace{-2mm}
In this paper, we propose \OURS, a novel 3D Generative Design framework that leverages data-driven algorithms to automate the robot hardware design process. Experiments demonstrate that the proposed algorithm can generate finger shapes for a target object that achieves more stable and robust grasps. 
However, since our algorithm focuses on geometry optimization, it does not optimize for the gripper strength, kinematic structure and grasping policy. The algorithm also does not explicitly reason about objects' physical material properties such as their friction and deformation. As future directions, it will be interesting to investigate the joint optimization of gripper geometry, kinematic structure, material, and grasping policy. We would also like to study how to apply this generative design approach to other design tasks beyond robot gripper such as furniture  or tool design, where the object's functionality is strongly influenced by their 3D geometry.

\acknowledgments{
We would like to thank Alberto Rodriguez, Hod Lipson, and Zhenjia Xu for fruitful discussions, and Google for the UR5 robot hardware.  This work was supported in part by the Amazon Research Award, the Columbia School of Engineering, as well as the National Science Foundation under CMMI-2037101. 
}
{\footnotesize
\bibliography{root}
}
\appendix


\section{Justification on Sim2Real transfer}
We have considered the following aspects to ensure that the algorithm can be applied to the real-world robotics systems:

\textbf{Manufacturing constraints:}
In our experiments, we give a score of zero to the generated fingers which are not feasible, i.e., the largest connected component of the volume is not significantly connected to the base. We were able to get $100 \%$ feasible fingers on the test set, showing that all the fingers can be manufactured.

\textbf{Robustness against perturbations to object orientation}
One of the major problems in Sim2Real transfer is the imprecise placement of objects. We specifically included robustness to perturbations in object orientation as one of our design objectives (in paper Sec. 3). We have seen in experiments that the generated fingers can achieve an average robustness of $78.8 \%$ (in paper Tab. 1), showing that the generated fingers are robust against perturbations in object orientation. 

\textbf{Stable against the application of external forces}
Another major challenge in Sim2Real transfer is the presence of unwanted forces on the object after it is grasped. These forces may arise due to jerky movement or unwanted shakes of the robotic arm carrying gripper or due to collision with other tools or objects in the environment. To mitigate this, we have specifically included force stability as one of our design objectives (in paper Sec. 3) where we apply external forces in the lower hemisphere of the object after it is grasped. We have seen in experiments that the generated fingers have average stability of $63.2\%$, which demonstrates that Fit2Form's grippers are likely to maintain its grasp through such unexpected external forces.

\textbf{Generating target object representation using accurate depth images:}
Input to our approach are TSDF volumes of the target objects, which are computed from multi-view depth images instead of directly voxelizing the 3D object meshes. This procedure closely resembles the sensory input that can be obtained by the real-world system (e.g., RGB-D cameras like Intel RealSense).  Therefore, our approach is likely to succeed with real-world data.

\textbf{Simulation of dynamic grasping process:}
We simulate the entire dynamic grasping process with contact simulation using the PyBullet physics engine \cite{coumans2019} for our experiments, which allows the target to be moved as the gripper closes, during lifting and application of external forces. This dynamic interaction in this simulation more closely mimics the real world in comparison to other grasp synthesis works \cite{goins2016implementation} which assumes the object to be static during grasp. We believe that it adds significantly to the chances of a successful Sim2Real transfer.

\textbf{Testing our approach and baselines with uncertainties in various system parameters:}
Apart from the above considerations in our approach, we have also tested our approach by varying system parameters which are most likely to have uncertainties in the real world. They are: (a) [Original] original parameters, i.e., object mass $50$g, object lateral friction $0.2$, object placement x $0$, y $0$ cm, (b) [m $100$g] object mass $100$g, (c) [m $150$g] object mass $150$g, (d) [$\mu$ $0.1$] object lateral friction $0.1$ (e) [$\mu$ $0.05$] object lateral friction $0.05$ (f) [ x $\pm 0.5$, y$\pm1$cm] object position will be uniformly selected from $x \in \pm[0.1, 0.5]$, $y \in \pm[0.1, 1]$ cm. Here $x$ represents the gripper closing direction and have small uncertainty ($0.5$ cm) as compared to the y direction ($1$ cm) because a larger deviation in $x$ almost always causes the object to collide with gripper during the gripper's downward motion. (g) [ x $\pm 0.5$, y$\pm2$cm] object position will be uniformly selected from $x \in \pm[0.1, 0.5]$, $y \in \pm[0.1, 2]$ cm. 

\begin{table}[t]
\centering
\small
\setlength\tabcolsep{4pt}
\begin{tabular}{lccccccc}
\toprule
         &  & \multicolumn{4}{c}{Object properties}  & \multicolumn{2}{c}{Postion Perturbation} \\  
 \cmidrule(lr){3-6} \cmidrule(lr){7-8} 
         & Original & m $100$g & m $150$g & $\mu$ $0.1$ & $\mu$ $0.05$ & x $\pm 0.5$, y$\pm1$cm & x$\pm0.5$, y$\pm2$cm \\
\midrule
WSG50    & 25.8     & 2.2   & 1.5   & 6.7    & 2.0    & 24.8          & 21.8         \\
Imprint  & 78       & 57.9  & 49.4  & 68.4   & 63.9   & 75.6          & 72.8        \\
Fit2Form & \textbf{88.9}     & \textbf{63.9}  & \textbf{51.6}  & \textbf{76.3}  & \textbf{73.0}   & \textbf{86.3}          & \textbf{83.6}        \\
\bottomrule
\end{tabular}
\vspace{1mm}
\caption{Testing our approach and baselines with uncertainties in various system parameters}
\label{tab:pert_results}
\vspace{-5mm}
\end{table}

As we can see from Tab. \ref{tab:pert_results}, Fit2Form can outperform all baselines even with strong perturbations to the desired system parameters.

\section{Additional Details}
\subsection{Imprint-based gripper}
\vspace{-2mm}
\begin{wrapfigure}{R}{0.48\textwidth}
    \centering
    \vspace{-3mm}
    \includegraphics[width=\linewidth]{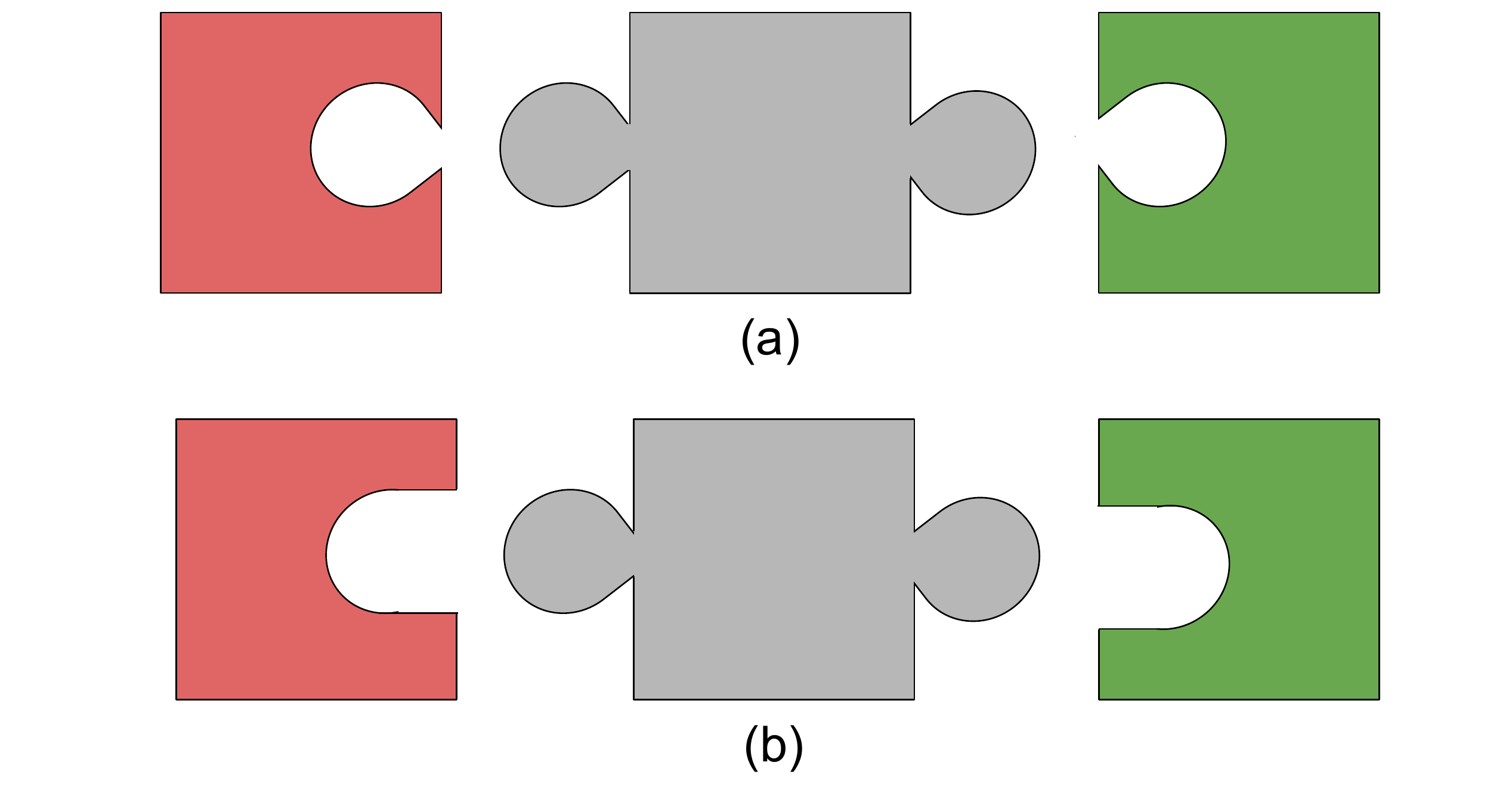} \vspace{-2mm}
    \caption{ \textbf{Imprint fingers} (a) Naively inverting the object geometry (b) Inverting the object geometry created by only two views along the grasp direction. }
    \label{fig:imprint}
    \vspace{-5mm}
\end{wrapfigure}
An intuitive way to create an imprint gripper for a given object is to use the profile created by the inverse of object geometry as fingers. Using the inverse object geometry provides maximum contact area and tightly cages the object. However, naively using the inverse object volume can cause the gripper to collide with the object prematurely and prevent it from tightly closing in on the object (Fig. \ref{fig:imprint}a). To generate imprint fingers for a given object, a TSDF volume of the object is created using only two views along the grasp direction. This volume is then negated and split along the mid-plane to give left and right imprint fingers. Using only two views mitigates the problem shown in Fig. \ref{fig:imprint}b and ensures that the resulting fingers can close properly.

\subsection{Network architecture}

In the following description, all convolution layers $C$ and convolution transpose layers $CT$ are represented by $C(n,k,s)$, where $n$ is the number of kernels, $k$ is the kernel size, and $s$ is the stride. $FC(n)$ represents fully connected layer with $n$ output nodes. All $C$, $CT$, $FC$ layers are followed by Batch Normalization \cite{ioffe2015batch} and Leaky ReLU activation, unless another activation is explicitly mentioned right after the layer. For the Generator network, padding size is always zero for both $C$ and $CT$ layers. Following is the description for three components of the Generator network:

Shape Encoder: $C(8,3,1)-C(16,3,1)-C(32,3,1)-C(64,3,1)-C(64,3,2)-C(32,3,1)-C(16,3,1)-C(8,3,1)-C(4,3,1)$

Shape Decoder: $FC(1024)-FC(2048)-FC(4096)-CT(64,3,1)-CT(64,3,1)-CT(32,3,3)-CT(16,3,1)-CT(8,3,1)-CT(4,3,1)-CT(2,3,1)-CT(1,3,1)-CT(1,3,1)-CT(1,3,1)-CT(1,3,1)-TanH$

Generator Embedding network:
$FC(2096)-FC(4096)-FC(1372)-TanH$

Apart from convolution layers, the Fitness network also uses residual blocks. A residual block is represented by $R(N)$ and is made up of $C(N,3,1)-ReLU-C(N,3,1)-ReLU$ where the input is added to the output of the second 3D convolution layer just before applying the ReLU activation. All $C$, $R$ layers in the Fitness Network has padding size $1$ except the first layer for which padding of size $2$ is used.

Fitness network: $C(32,5,2) - R(32) - R(32) - C(64, 3, 2) - R(64) - R(64) - C(128,3,2) - R(128)-B(256,3,2)-R(256) - C(512,3,2) - R(512) - C(1024,3,2) - R(1024) - C(1024,3,2) - FC(512) - FC(256) - FC(128) - FC(11) - Sigmoid$

The Fitness network and the Generator network are both optimized using the ADAM optimizer with weight decay $10^{-5}$ and learning rate $10^{-4}$ in all phases of the training.

\end{document}